# SHOP2: An HTN Planning System

**Dana Nau**     NAU@CS.UMD.EDU
*Dept. of Computer Science, and Institute for Systems Research*
*University of Maryland, College Park, MD 20742 USA*

**Tsz-Chiu Au**     CHIU@CS.UMD.EDU
*Dept. of Computer Science*
*University of Maryland, College Park, MD 20742 USA*

**Okhtay Ilghami**     OKHTAY@CS.UMD.EDU
*Dept. of Computer Science*
*University of Maryland, College Park, MD 20742 USA*

**Ugur Kuter**     UKUTER@CS.UMD.EDU
*Dept. of Computer Science*
*University of Maryland, College Park, MD 20742 USA*

**J. William Murdock**     MURDOCKJ@US.IBM.COM
*IBM Watson Research Center*
*19 Skyline Dr.*
*Hawthorne, NY 10532 USA*

**Dan Wu**     DANDAN@CS.UMD.EDU
*Dept. of Computer Science*
*University of Maryland, College Park, MD 20742 USA*

**Fusun Yaman**     FUSUN@CS.UMD.EDU
*Dept. of Computer Science*
*University of Maryland, College Park, MD 20742 USA*

## Abstract

The SHOP2 planning system received one of the awards for distinguished performance in the 2002 International Planning Competition. This paper describes the features of SHOP2 which enabled it to excel in the competition, especially those aspects of SHOP2 that deal with temporal and metric planning domains.

## 1. Introduction

SHOP2, Simple Hierarchical Ordered Planner 2 (Nau, Muñoz-Avila, Cao, Lotem, & Mitchell, 2001), is a domain-independent planning system based on Hierarchical Task Network (HTN) planning. In the 2002 International Planning Competition, SHOP2 received one of the top four awards, one of the two awards for distinguished performance. This paper describes some of the characteristics of SHOP2 that enabled it to excel in the competition.

Like its predecessor SHOP (Nau, Cao, & Muñoz-Avila, 1999), SHOP2 generates the steps of each plan in the same order that those steps will later be executed, so it knows the current state at each step of the planning process. This reduces the complexity of reasoning by eliminating a great deal of uncertainty about the world, thereby making it easy to incorporate substantial expressive power into the planning system. Like SHOP,

SHOP2 can do axiomatic inference, mixed symbolic/numeric computations, and calls to external programs.

SHOP2 also has capabilities that go significantly beyond those of SHOP:

- SHOP2 allows tasks and subtasks to be partially ordered; thus plans may interleave subtasks from different tasks. This often makes it possible to specify domain knowledge in a more intuitive manner than was possible in SHOP.

- SHOP2 incorporates many features from PDDL, such as quantifiers and conditional effects.

- If there are alternative ways to satisfy a method's precondition, SHOP2 can sort the alternatives according to a criterion specified in the definition of the method. This gives a convenient way for the author of a planning domain to tell SHOP2 which parts of the search space to explore first. In principle, such a technique could be used with any planner that plans forward from the initial state.

- So that SHOP2 can handle temporal planning domains, we have a way to translate temporal PDDL operators into SHOP2 operators that maintain bookkeeping information for multiple timelines within the current state. In principle, this technique could be used with any non-temporal planner that has sufficient expressive power.

The rest of this paper is organized as follows. Section 2 gives some background on HTN planning, and Section 3 describes SHOP2's features and planning algorithm. Section 4 describes how to write domain descriptions for SHOP2: in particular, Section 4.1 discusses basic problem-solving strategies, and Sections 4.2 and 4.3 describe aspects of SHOP2 that are specific to handling temporal and metric domain features. Section 5 discusses SHOP2's performance in the competition, Section 6 discusses related work, and Section 7 gives a summary and conclusion. Appendix A contains a SHOP2 domain description for one of the problem domains in the planning competition.

## 2. HTN Planning

HTN planning is like classical AI planning in that each state of the world is represented by a set of atoms, and each action corresponds to a deterministic state transition. However, HTN planners differ from classical AI planners in what they plan for, and how they plan for it.

The objective of an HTN planner is to produce a sequence of actions that perform some activity or *task*. The description of a planning domain includes a set of operators similar to those of classical planning, and also a set of *methods*, each of which is a prescription for how to decompose a task into *subtasks* (smaller tasks). Figure 1 gives a simple example. Given a planning domain, the description of a planning problem will contain an initial state like that of classical planning—but instead of a goal formula, the problem specification will contain a partially ordered set of tasks to accomplish.

Planning proceeds by using the methods to decompose tasks recursively into smaller and smaller subtasks, until the planner reaches *primitive tasks* that can be performed directly using the planning operators. For each nonprimitive task, the planner chooses an applicable method, instantiates it to decompose the task into subtasks, and then chooses and instantiates methods to decompose the subtasks even further, as illustrated in Figure 2. If the plan later turns out to be infeasible, the planning system will need to backtrack and try other methods.

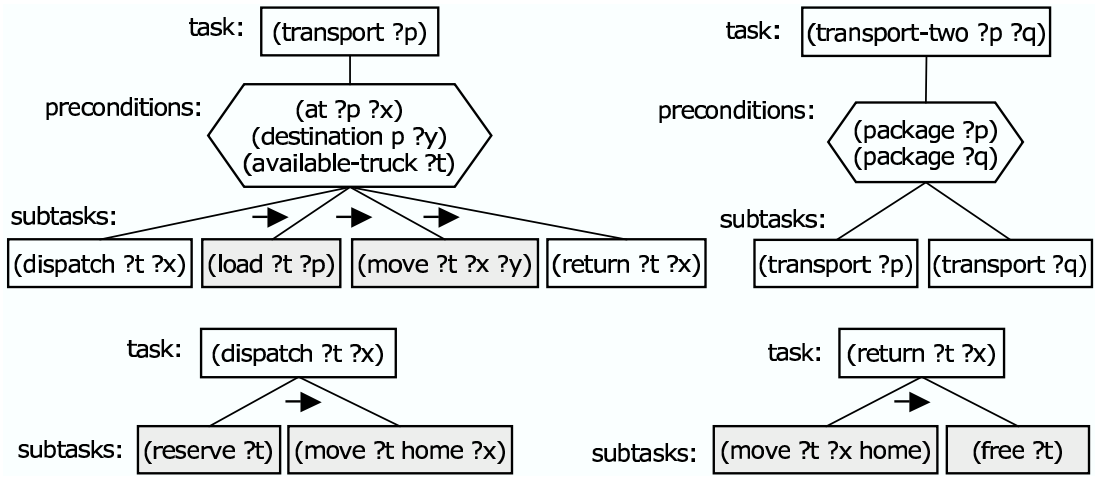

Figure 1: Methods for transporting a package ?p, transporting two packages ?p and ?q, dispatching a truck ?t, and returning the truck. Arrows are ordering constraints. The shaded subtasks are *primitive* tasks that are accomplished by the following planning operators: (load ?t ?p) loads ?p onto ?t; (move ?t ?x ?y) moves ?t from ?x to ?y; (reserve ?t) deletes (available-truck ?t) to signal that the truck is in use; (free ?t) adds (available-truck ?t) to signal that the truck is no longer in use.

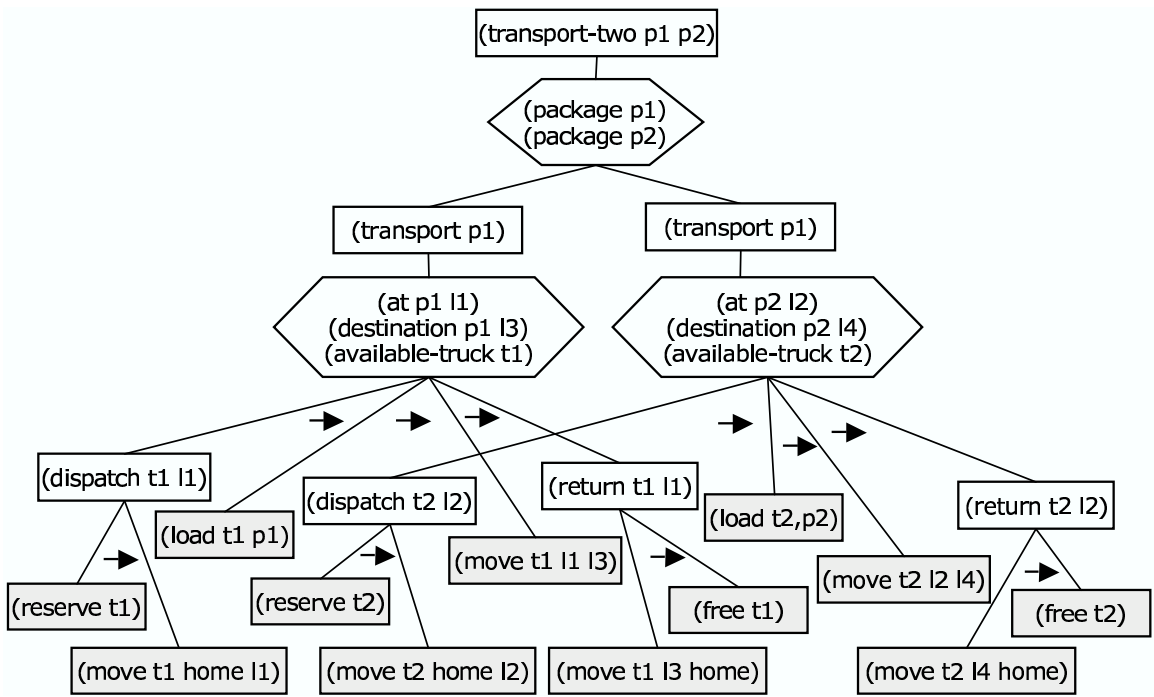

Figure 2: A plan for accomplishing (transport-two p1 p2) from the following initial state: {(package p1), (at p1 l1), (destination p1 l3), (available-truck t1), (at t1 home), (package p2), (at p2 l2), (destination p2 l4), (available-truck t2), (at t2 home)}.

HTN methods generally describe the "standard operating procedures" that one would normally use to perform tasks in some domain (e.g., see Figure 1) Most HTN practitioners would argue that such representations are more appropriate for many real-world domains than are classical planning operators, as they better characterize the way that users think about problems.

Like most other HTN planners, SHOP2 is "hand-tailorable:" its planning engine is domain-independent, but the HTN methods may be domain-specific, and the planner can be customized to work in different problem domains by giving it different sets of HTN methods. The ability to use domain-specific problem-solving knowledge can dramatically improve a planner's performance, and sometimes make the difference between solving a problem in exponential time and solving it in polynomial time (e.g., Gupta & Nau, 1992; Slaney & Thiébaux, 2001). In experimental studies (e.g., Nau et al., 1999, 2001; Bacchus & Kabanza, 2000), hand-tailorable planners have quickly solved planning problems orders of magnitude more complicated than those typically solved by "fully automated" planning systems in which the domain-specific knowledge consists only of the planning operators.

## 3. Features of SHOP2

This section describes SHOP2's planning algorithm and some of SHOP2's distinctive features.

### 3.1 Basic Elements of a Domain Description

A *domain description* is a description of a planning domain, consisting of a set of methods, operators, and axioms. Below we describe each of these briefly; additional details appear in Section 4.

#### 3.1.1 TASKS

A *task* represents an activity to perform. Syntactically, a task consists of a *task symbol* followed by a list of arguments. A task may be either *primitive* or *compound*. A primitive task is one that is supposed to be accomplished by a planning operator: the task symbol is the name of the planning operator to use, and the task's arguments are the parameters for the operator. A compound task is one that needs to be decomposed into smaller tasks using a method; any method whose head unifies with the task symbol and its arguments may potentially be applicable for decomposing the task. The details are discussed in the following subsections.

#### 3.1.2 OPERATORS

Each operator indicates how a primitive task can be performed. The operators are very similar to PDDL operators: each operator $o$ has a *head* head($o$) consisting of the operator's name and a list of parameters, a *precondition* expression pre($o$) indicating what should be

```
(:method
  ; head
      (transport-person ?p ?c2)
  ; precondition
      (and
         (at ?p ?c1)
         (aircraft ?a)
         (at ?a ?c3)
         (different ?c1 ?c3))
  ; subtasks
      (:ordered
         (move-aircraft ?a ?c1)
         (board ?p ?a ?c1)
         (move-aircraft ?a ?c2)
         (debark ?p ?a ?c2)))
```

Figure 3: A SHOP2 method for a simplified version of the ZenoTravel domain.

true in the current state in order for the operator to be applicable, and a *delete list* del(*o*) and *add list* add(*o*) giving the operator's negative and positive effects. Like in PDDL, the preconditions and effects may include logical connectives and quantifiers. The operators also can do numeric computations and assignments to local variables (an example appears later in Figure 11). Just as in PDDL, no two operators can have the same name; thus for each primitive task, all applicable actions are instances of the same operator.

Each operator also has an optional *cost* expression (the default value is 1). This expression can be arbitrarily complicated and can use any of the variables that appear in the operator's head and precondition. The cost of a plan is the sum of the costs of the operator instances.

### 3.1.3 Methods

Each method indicates how to decompose a compound task into a partially ordered set of subtasks, each of which can be compound or primitive. The simplest version of a method has three parts: the *task* for which the method is to be used, the *precondition* that the current state must satisfy in order for the method to be applicable, and the *subtasks* that need to be accomplished in order to accomplish that task.

As an example, Figure 3 is a simplified version of a SHOP2 method for one of the domains in the AIPS-2002 Planning Competition, the ZenoTravel domain. This method gives a way to transport a person ?p by aircraft from one location ?c1 to another location ?c2 if the aircraft is not already at ?c1.[1] The :ordered keyword specifies that the subtasks are totally ordered: first move the aircraft to ?c1, then board the person, then move the aircraft to ?c2, then debark the person.[2] To specify an unordered set of subtasks, we would

---

1. Any symbol that begins with a question mark is a variable name.
2. The method in the figure would have the same meaning if :ordered were omitted. If the list of subtasks does not begin with :ordered or :unordered, SHOP2 assumes :ordered.

```
(:-
  ; head
    (enough-fuel ?plane ?current-position ?destination ?speed)
  ; tail
    (and (distance ?current-position ?destination ?dist)
         (fuel ?plane ?fuel-level)
         (fuel-burn ?speed ?rate)
         (eval (>= ?fuel-level (* ?rate ?dist)))))
```

Figure 4: A SHOP2 axiom for a simplified version of the ZenoTravel domain.

use the keyword :unordered rather than :ordered; more complicated partial orderings can be specified using nested combinations of :ordered and :unordered.[3]

More generally, a method $m$ may have the form

$$(\text{:method head}(m)\ p_1\ t_1\ p_2\ t_2\ \ldots),$$

where head($m$) is a task called the *head* of $m$, each $p_i$ is a precondition expression and each $t_i$ is a partially ordered set of subtasks. The meaning of this is analogous to an if-then-else: it tells SHOP2 that if $p_1$ is satisfied then $t_1$ should be used, otherwise if $p_2$ is satisfied then $t_2$ should be used, and so forth. To keep the descriptions in this paper simple, we will assume without loss of generality that there is only one precondition expression pre($m$) and one set of subtasks sub($m$).

In general, there may be several alternative ways of accomplishing head($m$). There may be more than one method whose head is head($m$), more than one set of variable bindings that satisfy pre($m$), more than one ordering consistent with sub($m$), or more than one possible way to accomplish some of the subtasks in sub($m$). These alternatives produce branches in SHOP2's search space.

### 3.1.4 Axioms

The precondition of each method or operator may include conjunctions, disjunctions, negations, universal and existential quantifiers, implications, numerical computations, and external function calls. Furthermore, axioms can be used to infer preconditions that are not explicitly asserted in the current state. The axioms are generalized versions of Horn clauses, written in a Lisp-like syntax: for example, (:- *head tail*) says that *head* is true if *tail* is true. The tail of the clause may contain anything that may appear in the precondition of an operator or method.

As an example, the axiom shown in Figure 4 says that a plane has enough fuel to reach ?destination if the following conditions are satisfied: the distance to travel is ?dist, the fuel level is ?fuel-level, the burn rate is ?rate, and ?fuel-level is not less than the product of ?rate and ?distance. The last of these conditions is handled using an external function call, as described below.

---
3. This notation does not allow every possible possible partial ordering, but that has not been a problem in practice; and this notation is less clumsy than those that allow every possible partial ordering.

```
procedure SHOP2(s, T, D)
    P = the empty plan
    T_0 ← {t ∈ T : no other task in T is constrained to precede t}
    loop
        if T = ∅ then return P
        nondeterministically choose any t ∈ T_0
        if t is a primitive task then
            A ← {(a, θ) : a is a ground instance of an operator in D, θ is a substi-
                    tution that unifies {head(a), t}, and s satisfies a's preconditions}
            if A = ∅ then return failure
            nondeterministically choose a pair (a, θ) ∈ A
            modify s by deleting del(a) and adding add(a)
            append a to P
            modify T by removing t and applying θ
            T_0 ← {t ∈ T : no task in T is constrained to precede t}
        else
            M ← {(m, θ) : m is an instance of a method in D, θ unifies {head(m), t},
                    pre(m) is true in s, and m and θ are as general as possible}
            if M = ∅ then return failure
            nondeterministically choose a pair (m, θ) ∈ M
            modify T by removing t, adding sub(m), constraining each task
                in sub(m) to precede the tasks that t preceded, and applying θ
            if sub(m) ≠ ∅ then
                T_0 ← {t ∈ sub(m) : no task in T is constrained to precede t}
            else T_0 ← {t ∈ T : no task in T is constrained to precede t}
    repeat
end SHOP2
```

Figure 5: A simplified version of the SHOP2 planning procedure.

If the tail of a clause (or the precondition of an operator or method) contains a negation, it is handled in the same way as in Prolog: the theorem prover takes (not $a$) to be true if it cannot prove $a$.

### 3.1.5 External Function Calls

External function calls are useful, for example, to do numeric evaluations (e.g., in the ZenoTravel domain, to check the requirement that the available fuel must be greater than or equal to the product of the burn rate and the distance to be traveled). For example, in the competition, SHOP2 used a graph-algorithm library to compute the shortest paths in a graph. In principle, it would be possible to implement the graph algorithms as a set of methods. However, writing them as external functions allows them to run faster, and also makes it possible to access predefined code libraries.

### 3.2 The SHOP2 Algorithm

Figure 5 shows a simplified version of the SHOP2 planning procedure. The arguments include the initial state $s$, a partially ordered set of tasks $T$, and a domain description $D$.

As we mentioned earlier, SHOP2 plans for tasks in the same order that they will be executed. In order to do this, it nondeterministically chooses a task $t \in T$ that has no predecessors; $t$ is the first task that SHOP2 will start working on. At this point, there are two cases.

The first case is if $t$ is *primitive*, i.e., if $t$ can be accomplished directly using an action (i.e., an instance of a planning operator). In this case, SHOP2 finds an action $a$ that matches $t$ and whose preconditions are satisfied in $s$, and applies $a$ to $s$ (if no such action exists, then this branch of the search space fails).

The second case is where $t$ is *compound*, i.e., a method needs to be applied to $t$ to decompose it into subtasks. In this case, SHOP2 nondeterministically chooses a method instance $m$ that will decompose $t$ into subtasks (if no such method instance exists, then this branch of the search space fails).

If there is a solution plan that involves $m$, then the actions in $P$ will be the leaf nodes of a decomposition tree $D_P$ such as the tree shown in Figure 2. The precondition formula $\text{pre}(m)$ must be true in the state that immediately precedes the first action $a$ in $D_P$ that is a descendant of $m$. In order to ensure that $\text{pre}(m)$ is true in the correct state, SHOP2 needs to generate the leftmost branch of $D$ all the way down to the bottom, and evaluate $\text{pre}(m)$ in the state just before $a$. The last three lines of the loop ensure that this will happen, by telling SHOP2 that if the current method $m$ has any subtasks, SHOP2 should generate one of those subtasks before generating any other subtasks in the task network.

For example, SHOP2 could begin generating the plan in Figure 2 by first decomposing (transport-two p1 p2) into (transport p1) and (transport p2), and then nondeterministically choosing to decompose (transport p1) into {(dispatch t1 l1), (pickup t1 p1), (move t1 l1 l3)}. Having done that, SHOP2 would be required to decompose (dispatch t1 l11) before decomposing (transport p2), in order to guarantee that (dispatch t1 l1) and (reserve t1) occur in the same state of the world in which (available t1) was evaluated. The operator for (reserve t1) makes t1 unavailable, thus ensuring that when (transport p2) is decomposed later, the decomposition will use truck t2 rather than t1.

### 3.3 Additional Features

SHOP2 has several additional features in addition to the basic ones described earlier. This section describes the most significant ones.

#### 3.3.1 Sorting the Variable Bindings

When SHOP2 evaluates a method's precondition, it gets a list of all of the possible sets of variable bindings that satisfy the expression in the current state. Each set of variable bindings can lead to a different branch in SHOP2's search tree. This nondeterministic choice is implemented in SHOP2 via depth-first backtracking. For SHOP2 to find a good solution and to find it quickly, it is important to decide which set of variable bindings to try first.

For this purpose, SHOP2 has a "sort-by" construct that sorts the list of variable bindings by a specified criterion. This is especially useful when the planning problem is an

```
(:method
  ; head
      (transport-person ?p ?c2)
  ; precondition
      (:sort-by ?cost #'<
        (and (at ?p ?c1)
            (aircraft ?a)
            (at ?a ?c3)
            (different ?c1 ?c3)
            (cost-of ?a ?c3 ?c1 ?cost)))
  ; subtasks
      ((move-aircraft ?a ?c1)
        (board ?p ?a ?c1)
        (move-aircraft ?a ?c2)
        (debark ?p ?a ?c2)))
```

Figure 6: Using sort-by in a SHOP2 method for the simplified ZenoTravel domain.

optimization problem, e.g., a problem in which the objective is to find a plan having the least possible cost. With the sort-by construct, we can write a heuristic function to estimate the anticipated cost of each set of variable bindings, and sort the sets of variable bindings according to their heuristic-function values so that SHOP2 will try most promising one first.

For example, if we have the precondition

$$(\text{and (at ?here) (distance ?here ?there ?d)})$$

then there may be several different combinations of ?here, ?there, and ?d that satisfy this precondition. The expression

$$(\text{:sort-by ?d \#'> (and (at ?here) (distance ?here ?there ?d))})$$

will cause SHOP2 to consider the variable bindings in decreasing order of the value of ?d.

As a more complicated example, recall the precondition of the method in Figure 3. There may be several sets of variable bindings that satisfy this precondition in the current state. The reformulation of the precondition in Figure 6 tells SHOP2 to sort the sets of variable bindings in increasing order of the ?cost variable. This way, SHOP2 will look first at the alternative that has the lowest ?cost value.

### 3.3.2 Branch-and-Bound Optimization

SHOP2 allows the option of using branch-and-bound optimization to search for a least-cost plan. This option generally results in spending additional planning time in order to search for plans of superior quality. When using the branch-and-bound option, one can also specify a time limit for the search. If the search takes longer than the time limit, SHOP2 terminates the search and returns the best plan it has found so far; this functionality was partly inspired by anytime algorithms (Boddy & Dean, 1989).

### 3.3.3 PDDL Operator Translation

SHOP2's planning procedure can be proved to be sound and complete across a large set of planning problems, in the sense that if a set of methods and operators is capable of generating a solution for some problem, then the planning procedure is guaranteed to generate a correct plan (Nau et al., 2001). However, while such a proof tells us that the planning algorithm should work correctly if the domain description is correct, it does not tell us whether our domain description represents the same planning domain as a given set of PDDL planning operators.

In the AIPS-2000 planning competition, that problem caused difficulty for SHOP2's predecessor SHOP. The SHOP team was developing domain descriptions for SHOP purely by hand, and made some mistakes in writing two of the domains. Thus SHOP found incorrect solutions for some of the problems in those domains, so the judges disqualified SHOP from those domains.

While developing SHOP2, we wrote a translator program to translate PDDL operators into SHOP2 domain descriptions. The domain descriptions produced by the translator program are not sufficient for efficient planning with SHOP2: they need to be modified by hand in order to put in the domain knowledge, as described in Section 4. However, the translator program can at least provide a correct starting point.

### 3.3.4 Debugging Facilities

SHOP2 also includes several debugging facilities. The most important of these is a tracing mechanism: one can tell SHOP2 to trace any set of operators, methods, and axioms. For example, in Figure 8, we have given names (namely Case1 and Case2) to the two different clauses of a method. We can tell SHOP2 to trace either of these clauses or both of them; SHOP2 will print messages each time it enters and exits a clause that is being traced. Depending on the particular tracing options that one selects, the messages may include things such as the argument list, the current state of the world, and information about whether the operator, method or axiom succeeds or fails.

### 3.3.5 Protected Conditions and Anti-Interleaving

SHOP2's planning operators include a way to specify protected conditions. This feature is described briefly in Nau et al. (2001), but we will not bother to describe it here because we did not use it during the planning competition. In cases where we wanted to protect conditions from possible threats, we found it more convenient either to make use of flags similar to the available-truck flag in Figure 1, or to use the following "anti-interleaving" feature of SHOP2.

If a method $m$ has subtasks $t_1, \ldots, t_k$, and if any $t_i$ begins with the keyword :immediate, this tells SHOP2 that it should plan to do $t_i$ immediately after $t_{i-1}$ finishes, without trying to interleave other tasks between $t_{i-1}$ and $t_i$. Several examples of this appear in the appendix.

> task for each person: transport the person to his/her destination
>     (these tasks are unordered; thus their subtasks may be interleaved)
>
> task for each plane: transport the plane to its destination
>     (these tasks are unordered; thus their subtasks may be interleaved)
>
> method for transporting a person:
>     if the person is already at their desired destination then do nothing
>     else
>         select a plane
>         if the plane is not at the person's current position then move it there
>         hold the plane at the current position
>         board the person onto the plane
>         move the plane to the destination
>         debark the person at the destination
>
> method for transporting a plane:
>     if the plane is already at its desired destination then do nothing
>     else move the plane to the destination

Figure 7: Abstract tasks and methods for a simplified version of the ZenoTravel domain.

## 4. Developing Domain Descriptions for SHOP2

### 4.1 Basics

The first step in developing a domain description for SHOP2 is to formulate some abstract tasks and methods that constitute a reasonable problem-solving strategy. As an example, we will use a simplified version of ZenoTravel (one of the domains in the planning competition). The problem is to transport people from their current locations to their destination, by the use of any available airplanes. Figure 7 shows a set of abstract tasks and methods for transporting people and moving airplanes.

Once we have an abstract strategy like the one in Figure 7, we can implement it as a SHOP2 domain description consisting of methods, operators and axioms. For example, the method for transporting a person is shown in Figure 8. There may be more than one value of ?p that satisfies the precondition (plane ?p). If so, then which plane to use will be a nondeterministic branching point for SHOP2. At nondeterministic branching points, the domain description may include some heuristics to guide SHOP2's search; Section 4.3 discusses some of the ways to write such heuristics.

In the method for the transport-with-plane task, one of the preconditions will be whether a plane's fuel level will be enough to get the plane from its current position to its destination. Figure 4 shows an axiom for this precondition.

Actions, such as boarding people onto planes, debarking them from planes, and refueling the planes, can be modeled as operators in SHOP2. For example, the SHOP2 operator for boarding is given in Figure 9.

Nau, Au, Ilghami, Kuter, Murdock, Wu, & Yaman```
(:method
    ; head
        (transport-person ?person ?destination)
  Case1 ; a label for use in debugging
    ; preconditions
        (and (at ?person ?current-position)
             (same ?current-position ?destination))
    ; subtasks
        ()
  Case2 ; a label for use in debugging
    ; preconditions
        (and (at ?person ?current-position)
             (plane ?p))
    ; subtasks
        ((transport-with-plane ?person ?p ?destination)))
```

Figure 8: A SHOP2 implementation of one of the methods in Figure 7.

```
(:operator
  ; head
        (board ?person ?plane)
  ; preconditions
        (and (at ?person ?place) (at ?plane ?place))
  ; delete list
        ((at ?person ?place))
  ; add list
        ((in ?person ?plane)))
```

Figure 9: A SHOP2 operator for the simplified ZenoTravel domain.

### 4.2 Writing Temporal Domains

SHOP2's operators are at least as expressive as Level 2 actions in PDDL, but SHOP2 does not explicitly support the durative actions in Level 3 of PDDL, nor does SHOP2 have an explicit mechanism for reasoning about durative and concurrent actions.

However, SHOP2 still has enough expressive power to represent durative and concurrent actions, because it knows the current state at each step of the planning process and since its operators can assign values to variables and can do numeric calculations. This has allowed us to develop a preprocessing technique that we call *Multi-Timeline Preprocessing* (MTP). MTP is a technique for translating PDDL operators into SHOP2 operators that keep track of temporal information in the current state.

The pseudocode in Figure 10 is an algorithmic description of what MTP does. In principle, MTP could be automated—but in practice, we have always done it by hand, because it only needs to be done once for each planning domain.

390

SHOP2: An HTN Planning System> for every operator $o$ in the planning domain
>   add two parameters *?start* and *?duration* to $o$
>   in $o$'s precondition
>     add an assignment ?duration $\leftarrow d$
>       where $d$ is a formula for calculating $o$'s duration
>     add an assignment ?start $\leftarrow s$
>       where $s$ is a formula that takes the maximum of the write times of
>       all dynamic properties in $o$'s precondition and the read times of all
>       dynamic properties in $o$'s effects
>   for each dynamic property $p$ in $o$'s effects
>     add effects to change the value of *write-time(p)* to *?start + ?duration*
>   for each dynamic property that appears in $o$
>     add effects to change *read-time(p)* to the maximum
>       of *read-time(p)* and *?start + ?duration*

Figure 10: Multi-timeline preprocessing (MTP).

To keep the description of MTP simple, let us suppose that in each state $s$, every atom (p $c_1$ ... $c_n$) represents a single-valued property, i.e., there is at most one $c_n$ such that (p $c_1$ ... $c_{n-1}$ $c_n$) is true in $s$. A property is *dynamic* if an operator may change the value of $c_n$. For example, if the initial state contains (at plane1 city1) but if there is an operator that moves plane1 to a different location, then the location of plane1 is dynamic.

For each property $p$ that changes over time, MTP modifies the operators to keep track, within the current state, of the times at which the property changes and the times at which various preconditions depend on the property. The idea is that for each dynamic property $p$, the current state will contain two time-stamps: *read-time(p)*, which is the last time that any action read the value of $p$, and *write-time(p)*, which is the last time that any action modified the value of $p$. MTP modifies the operators in such a way that whenever an operator reads (i.e., accesses) a dynamic property, the operator will update the property's read-time, and if an operator writes (i.e., modifies) a dynamic property, it will update the property's write-time. Thus, instead of a single "global time," the current state will contain many "local times," namely a read-time and a write-time for each dynamic property.

MTP also inserts preconditions into each action to ensure that the action begins on or after the *read-time* of each property that it writes and the *write-time* of each property that it reads. This prevents two actions from overlapping in time if one of them writes to a property and the other reads from it. For example, a boarding operator and a fly operator on the same plane may not overlap, because the boarding operator requires that the plane be located in a particular city and the fly operator changes the location of the plane.

Figure 11 shows one of the SHOP2 operators produced by MTP for the ZenoTravel domain. The operator involves two dynamic properties: a vehicle's fuel level and its location. The operator reads both of these properties, so it may not start before their write times. However, it only writes one of them (the fuel level), so it may start before the read time of the vehicle's location. Thus refueling may be performed concurrently with any other actions that depend on the vehicle's location, but cannot be performed concurrently with any actions that modify the fuel level or that modify the vehicle's location.

391



```
(:operator (!refuel ?plane ?city ?start ?duration)
  ; preconditions
    ((aircraft ?plane)
      (city ?city)
      (at ?plane ?city)
      (fuel ?plane ?fuel-level)
      (capacity ?plane ?fuel-cap)
      (refuel-rate ?plane ?rate)
      (assign ?duration (/ (- ?fuel-cap ?fuel-level) ?rate))
      (write-time fuel ?plane ?t1)
      (write-time at ?plane ?t2)
      (read-time fuel ?plane ?t3)
      (assign ?start (eval (max ?t1 ?t2 ?t3)))
      (assign ?end (eval (+ ?start ?duration)))
      (read-time at ?plane ?t4)
      (assign ?new-value (eval (max ?t4 ?end))))
  ; delete list
    ((fuel ?plane ?fuel-level)
      (write-time fuel ?plane ?t1)
      (read-time fuel ?plane ?t3)
      (read-time at ?plane ?t4))
  ; add list
    ((fuel ?plane ?fuel-cap)
      (write-time fuel ?plane ?end)
      (read-time fuel ?plane ?end)
      (read-time at ?plane ?new-value)))
```

Figure 11: A sample SHOP2 operator produced by MTP.

### 4.3 Writing Domains that Include Optimization

In previous planning competitions, the planning benchmarks compared only the speed of the planners and the length of the output plans, so domain designers concentrated on trying to find a reasonably short plan as quickly as possible. In contrast, most of the problems in this year's competition included a linear objective function that needed to be optimized: the best plan was no longer the one that minimizes the number of steps, but instead was the one that minimized the objective-function value. We tried three approaches for searching for optimal plans:

1. Structure the SHOP2 methods in such a way as to take SHOP2 more-or-less directly toward a plan that minimizes the objective function.

2. Write methods, operators, and axioms to generate plans quickly, and use the "sort-by" feature to tell SHOP2 to sort the alternatives and try the most promising ones first.

3. Assign costs to the operators, and use a branch-and-bound search to find the best plan within the execution time limit.





The first approach works well if it is easy to tell which alternative will be best at each node of the search space. For example, if you know that in all problem instances the objective will be to minimize the total fuel used, then a perfect heuristic for the ZenoTravel domain is to always use the fly action instead of the zoom action. However, this approach doesn't work so well if it isn't immediately obvious which alternative is best. For example, if the objective is to minimize the total time, then a naive approach would be always to use the zoom action rather than the fly action, since the zoom action is faster. However, the zoom action is not always the best choice, because it requires more fuel and thus can cause delays for refueling.

The second approach is an extension of the first approach. Consider again the example in ZenoTravel domain where the objective is to minimize the total fuel used. In addition to making the planes fly instead of zoom; using a closer plane to transport a person also reduces the total fuel used. We can set this preference using the sort-by feature of SHOP2. In the precondition of the method for transporting a person we can sort the available planes according to the fuel they will use in order to pick up this person. This is a greedy approach. At each decision point we can sort the alternatives by cost, and go with the alternative that has the lowest objective-function value. Thus, this approach is not guaranteed to find the optimal solution. However, if combined with suitable heuristics, this approach results in near-optimal plans. In the competition we used this technique extensively, and it produced satisfactory plans even for the largest problems.

The third approach makes use of branch-and-bound optimization, as explained in Section 3.3.2. The main idea is to quickly define methods that will let you find a plan which may be poor in quality and then let SHOP2 perform branch-and-bound search in the plan space to find the least cost plan or the best plan it can find within the execution time limit.

For the third approach, there is a challenge in setting up the cost of each operator. For example, if the objective function requires minimizing the total time, then in order to take concurrency into account, the cost of an action $a$ should not always be equal to its duration. For example, suppose the latest event in the current partial plan is for a plane to arrive at an airport at time $t$, and two passengers need to board the plane. Then we need to add two boarding actions to the plan. Recall that in the ZenoTravel domain, all boarding actions take the same amount of time, $t_b$, and can be performed concurrently. However, SHOP2 needs to add the actions to the plan one at a time. The first boarding action will increase the total time by $t_b$, so its cost is $t_b$. However, the second boarding action will not increase the total time of the plan, so its cost is 0. Now, suppose we add a "refuel" action to the plan. This action can be done concurrently with the boarding actions, so its cost is $\max(0, t_r - t_b)$, where $t_r$ is the time needed to refuel.

It is possible to combine two or more of the above approaches. However, in our experience, using optimization (the third approach) did not provide much benefit for domain descriptions that already included the other two approaches. In these situations, SHOP2 would frequently find an optimal or nearly optimal plan even without optimization, which meant that the additional amount of time needed by branch-and-bound optimization would produce little or no benefit. Branch-and-bound optimization would perhaps be more useful in planning domains where the cost of a plan is something other than the sum of the costs of the operators; however, such domains did not occur in the planning competition.





In the International Planning Competition, we did not use the optimization approach in any official competition trial. For all of the competition domains, in our preliminary testing of SHOP2 with optimization and no time limits, SHOP2 was unable to find solutions within the amount of time that we were willing to let it run, except on the very smallest problems. One way to overcome this difficulty would have been to use time limits, but in our preliminary tests, this never provided significant improvements in cost across an entire problem set. One reason for this lack of improvement was that we spent a great deal of effort crafting the methods used in the competition. We think the third approach would be more useful in cases where it is not immediately clear how to implement the first two approaches, and one does not want to spend too much time devising a sophisticated domain description.

## 5. Competition Results

Fourteen planning systems competed in the 2002 International Planning Competition. SHOP2 received a "distinguished performance" award, one of the top four awards.

SHOP2 (along with TLPlan and TALPlanner) was one of three planners that solved problems in both the "hand-tailored" and "fully automated" tracks. SHOP2 was able to solve problems in the Strips, Numeric, HardNumeric, SimpleTime, Time, and Complex domains. SHOP2 solved more problems than any other planner in the competition: it solved 899 out of 904 problems, for a 99% success ratio.

Of the other two hand-tailorable planners, TLPlan solved 894 problems, nearly as many as SHOP2. Since TALPlanner didn't do numeric domains, it solved only 610 problems, but that still was several hundred more problems than the fully-automated planners solved.

In general, SHOP2 tended to be slower than TALPlanner and TLPlanner, although there was one domain (Satellite-HardNumeric) where SHOP2 was consistently the fastest. The speeds of the three hand-tailorable planners generally appeared to be polynomially related to each other, probably because these planners' domain knowledge enabled them to find solutions without doing very much backtracking. All three hand-tailorable planners were generally much faster than most of the fully-automated planners.

None of the three hand-tailorable planners dominated the other two in terms of plan quality. For each of them, there were situations where its solutions were significantly better or significantly worse than the other two.

## 6. Related Work

The following subsections discuss HTN planning, ordered task decomposition, and the other hand-tailorable planners that participated in the competition.

### 6.1 HTN Planning

HTN planning was first developed more than 25 years ago (Sacerdoti, 1990; Tate, 1977). Historically, most of the HTN-planning researchers have focused on practical applications. Examples include production-line scheduling (Wilkins, 1988), crisis management and logistics (Currie & Tate, 1991; Tate, Drabble, & Kirby, 1994; Biundo & Schattenberg, 2001), planning and scheduling for spacecraft (Aarup, Arentoft, Parrod, Stader, & Stokes, 1994;





Estlin, Chien, & Wang, 1997), equipment configuration (Agosta, 1995), manufacturability analysis (Hebbar, Smith, Minis, & Nau, 1996; Smith, Hebbar, Nau, & Minis, 1997), evacuation planning (Muñoz-Avila, Aha, Nau, Weber, Breslow, & Yaman, 2001), and the game of bridge (Smith, Nau, & Throop, 1998a, 1998b).

The development of a formal semantics for HTN planning (Erol, Nau, & Hendler, 1994; Erol, Hendler, & Nau, 1996) has shown that it is strictly more expressive than classical AI planning: there are some problems that can be expressed as HTN planning problems but not as classical planning problems.[4] Even if one places restrictions on HTN planning to restrict its expressive power to that of classical planning, it generally is much easier to translate classical planning problems into HTN planning problems than vice versa (Lotem, Nau, & Hendler, 1999).

### 6.2 Ordered Task Decomposition

Ordered task decomposition (Nau, Smith, & Erol, 1998) is a special case of HTN planning in which the planning algorithm always builds plans forward from the initial state of the world. In other words, an ordered-task-decomposition planner plans for tasks in the same order that the tasks will later be performed. The first applications of ordered task decomposition were tailor-made for specific application domains. The best known example is the code for declarer play that helped *Bridge Baron* win the 1997 world championship of computer bridge (Smith et al., 1998b).

SHOP2 is based on SHOP (Nau et al., 1999), a previous domain-independent ordered-task-decomposition planner that requires the subtasks of each method, and also the initial set of tasks for the planning problem, to be totally ordered rather than partially ordered. Thus in SHOP, subtasks of different tasks cannot be interleaved. SHOP2 extends SHOP by allowing the subtasks of each method to be partially ordered. Experiments have shown that this can allow SHOP2 to create plans more efficiently than SHOP, using domain descriptions simpler than those needed by SHOP (Nau et al., 2001). Both SHOP and SHOP2 are available as open-source software at ⟨http://www.cs.umd.edu/projects/shop⟩.

### 6.3 TLPlan and TALPlanner

Like SHOP2, TLPlan (Bacchus & Kabanza, 2000) and TALPlanner (Doherty & Kvarnström, 2001) competed in the "hand-tailored" track of the AIPS-2002 planning competition. TLPlan and TALPlanner are similar in many respects. Both of them do a forward-chaining search in which they apply planning operators to the current state to generate its successors. Thus, like SHOP2, they both know the current state of the world at every step of the planning process. To control their search, both planners use control rules that are written declaratively in temporal logic. These rules provide domain-specific knowledge to tell the planner which states are "bad" states, so that the planner can backtrack and try other

---

4. More specifically, HTN planning is Turing-complete: even undecidable problems can be expressed as HTN planning problems. It remains Turing-complete even if we restrict the tasks and the logical atoms to be purely propositional (i.e., to have no arguments at all). In contrast, classical planning only represents planning problems for which the solutions are regular sets. Planners such as TLPlan (Bacchus & Kabanza, 2000) and TALPlanner (Doherty & Kvarnström, 2001) overcome this limitation of classical planning by extending the formalism to include function symbols.





paths in the search space. One difference between TLPlan and TALPlanner is that TLPlan uses a linear modal tense logic, while TALPlanner use TAL, a narrative-based linear temporal logic used for reasoning about action and change in incompletely specified dynamic environments.

The main difference between both of these planners and SHOP2 is the kind of control knowledge they use: TLPlan and TALPlanner use their temporal formulas to tell which part of the search space should be avoided, whereas SHOP2 uses its HTN methods to tell which parts of the search space should be explored. SHOP2's search space consists only of those nodes that are reachable using its HTN methods, whereas TLPlan and TALPlanner can explore any part of the search space that avoids the "bad" states and their successors. It is hard to say which type of control knowledge is more effective. Bacchus and Kabanza (2000) argue that the two types are useful in different situations and that combining them is a useful topic for future research.

## 7. Summary and Conclusions

The primary difference between SHOP2 and most other HTN planners is that SHOP2 plans for tasks in the same order that they will be executed, and thus it knows the current state at each step of the planning process. This reduces the complexity of reasoning by removing a great deal of uncertainty about the world, which has made it easy for us to incorporate substantial expressive power into SHOP2. In addition to the usual HTN methods and operators, SHOP2's domain descriptions may include axioms, mixed symbolic/numeric conditions, and external function calls. The planning procedure is Turing-complete, and is sound and complete over a large class of planning problems (Nau et al., 2001).

Like other HTN planning systems, SHOP2 plans by decomposing tasks into subtasks. A key idea in using any HTN planner is to design a set of methods that encode something akin to "standard operating procedures" that capture multi-step techniques for refining a task. Some kinds of domain characteristics are much more natural to express in an HTN formalism than in action-based formalism; see Lotem et al. (1999) for a description of some of the issued involved.

As an example, consider the UM-Translog-2 domain, which we wrote for use as a problem domain in the AIPS-2002 planning competition (Wu & Nau, 2002). UM-Translog-2 is a straightforward generalization of the UM Translog domain (Andrews, Kettler, Erol, & Hendler, 1995); the generalizations include numeric information such as distances, fuel usage, and so forth. It was relatively straightforward to formulate UM-Translog-2 as an HTN planning domain. However, it was much more difficult to figure out how to formulate UM-Translog-2 as a PDDL domain for use by other competitors in the planning competition; that task took several months to accomplish.

As with most other HTN planning systems, SHOP2's development was originally motivated not by the planning competition but instead to try to solve practical planning problems. For example, JSHOP (a Java implementation of SHOP2's predecessor SHOP) is the generative-planning component of the HICAP system for planning evacuation operations (Muñoz-Avila et al., 2001), and we are currently incorporating SHOP2 into HICAP as a replacement for JSHOP. We are very pleased that SHOP2's capabilities also gave it the ability to excel in the International Planning Competition!






## Acknowledgments

This work was supported in part by the following grants, contracts, and awards: Air Force Research Laboratory F30602-00-2-0505, Army Research Laboratory DAAL0197K0135, Naval Research Laboratory N00173021G005, and the University of Maryland General Research Board. The opinions expressed in this paper are those of authors and do not necessarily reflect the opinions of the funders.

We also wish to thank the anonymous reviewers, whose comments helped us to make significant improvements to this paper.


## Appendix A. SHOP2 Domain Description for the ZenoTravel Domain (Numeric Version)

In the AIPS-2002 Planning Competition, there were four different versions of the ZenoTravel domain: the Strips version, the Numeric version, the Simple Time version, and the Time version. We developed SHOP2 domain descriptions for all four versions.

What follows is our domain description for the Numeric version of the ZenoTravel domain. The operators in our domain description are translated from the original PDDL coding using a rough approximation of the MTP process in Figure 10.

```
(defdomain ZENOTRAVEL
  (

  (:- (same ?x ?x) ())
  (:- (different ?x ?y) ((not (same ?x ?y))))

  (:-(possible-person-in ?city)
      ((person ?p) (at ?p ?city) (goal ?p ?city2)
       (different ?city2 ?city)))

  (:operator (!!cost ?end)
     ((maxtime ?max)
      (assign ?newmax (eval (if (< ?max ?end) ?end ?max))))
     ((maxtime ?max))
     ((maxtime ?newmax))
     (- ?newmax ?max))

  (:method (board ?p ?a ?c)
  ((write-time ?a ?start))
       ((!board ?p ?a ?c ?start 1)
   (:immediate !!cost (call + ?start 1))))

  (:operator (!board ?p ?a ?c ?start ?duration)
       ((person ?p) (aircraft ?a) (city ?c)
  (at ?a ?c) (at ?p ?c) (onboard ?a ?num)
  (read-time ?a ?pmax) (assign ?new-num (+ ?num 1))
       (assign ?newpmax (max ?pmax (+ ?start ?duration 0.01L0))))
       ((onboard ?a ?num) (read-time ?a ?pmax) (at ?p ?c) (dest ?a ?c))
       ((onboard ?a ?new-num) (read-time ?a ?newpmax) (in ?p ?a)))
```





```
 0.001)

 (:method (debark ?p ?a ?c)
      ((write-time ?a ?start))
      ((!debark ?p ?a ?c ?start 1)
 (:immediate !!cost (call + ?start 1))))

 (:operator (!debark ?p ?a ?c ?start ?duration)
    ((person ?p) (aircraft ?a) (city ?c)
     (at ?a ?c) (in ?p ?a) (onboard ?a ?num)
     (read-time ?a ?pmax) (assign ?new-num (- ?num 1))
     (assign ?newpmax (max ?pmax (+ ?start ?duration 0.01L0))))
     ((onboard ?a ?num) (read-time ?a ?pmax) (in ?p ?a) (dest ?a ?c))
     ((onboard ?a ?new-num) (read-time ?a ?newpmax) (at ?p ?c))
 0.001)

(:method (refuel ?a ?c)
        ((write-time ?a ?start) (read-time ?a ?pmax)
         (capacity ?a ?cap) (fuel ?a ?fuel)
    (eval (> ?cap ?fuel))
    (assign ?duration 1)
         (assign ?end (+ ?start ?duration 0.01L0))
         (assign ?newpmax (max ?pmax ?end)))
        ((!!ra ((read-time ?a ?pmax))
               ((read-time ?a ?newpmax)))
         (:immediate !refuel ?a ?c ?start ?duration)
         (:immediate !!cost ?end)))

(:operator (!refuel ?a ?c ?start ?duration)
    ((aircraft ?a) (city ?c) (at ?a ?c)
     (fuel ?a ?fuel) (capacity ?a ?cap))
    ((fuel ?a ?fuel))
    ((fuel ?a ?cap))
0.001)

(:method (zoom ?a ?c1 ?c2)
  ((write-time ?a ?astart) (read-time ?a ?pmax)
   (distance ?c1 ?c2 ?dist)
   (fuel ?a ?fuel) (fast-burn ?a ?burn)
   (eval (>= ?fuel (* ?dist ?burn)))
   (assign ?duration 1)
   (assign ?start (max ?pmax ?astart))
   (assign ?end (+ ?start ?duration 0.01L0)))
  ((!!ra ((write-time ?a ?astart) (read-time ?a ?pmax))
         ((read-time ?a 0) (write-time ?a ?end)))
   (:immediate !zoom ?a ?c1 ?c2 ?start ?duration)
   (:immediate !!cost ?end)))

(:operator (!zoom ?a ?c1 ?c2 ?start ?duration)
   ((aircraft ?a) (city ?c1) (city ?c2) (onboard ?a ?num)
    (zoom-limit ?a ?limit) (eval (<= ?num ?limit))
```





```
    (at ?a ?c1) (distance ?c1 ?c2 ?dist) (fast-burn ?a ?burn)
    (total-fuel-used ?total-fuel)
    (assign ?new-total (+ ?total-fuel (* ?dist ?burn)))
    (fuel ?a ?fuel)
    (assign ?new-fuel (- ?fuel (* ?dist ?burn))))
   ((at ?a ?c1) (total-fuel-used ?total-fuel) (fuel ?a ?fuel) )
   ((at ?a ?c2) (total-fuel-used ?new-total) (fuel ?a ?new-fuel))
 0.001)

(:method (fly ?a ?c1 ?c2)
  ((write-time ?a ?astart) (read-time ?a ?pmax)
    (distance ?c1 ?c2 ?dist)
    (fuel ?a ?fuel) (slow-burn ?a ?burn)
    (eval (>= ?fuel (* ?dist ?burn)))
    (assign ?duration 1)
    (assign ?start (max ?pmax ?astart))
    (assign ?end (+ ?start ?duration 0.01L0)))
   ((!!ra ((write-time ?a ?astart) (read-time ?a ?pmax))
         ((read-time ?a 0) (write-time ?a ?end)))
    (:immediate !fly ?a ?c1 ?c2 ?start ?duration)
    (:immediate !!cost ?end)))

(:operator (!fly ?a ?c1 ?c2 ?start ?duration)
   ((aircraft ?a) (city ?c1) (city ?c2)
    (at ?a ?c1) (distance ?c1 ?c2 ?dist) (slow-burn ?a ?burn)
    (total-fuel-used ?total-fuel)
    (assign ?new-total (+ ?total-fuel (* ?dist ?burn)))
    (fuel ?a ?fuel)
    (assign ?new-fuel (- ?fuel (* ?dist ?burn))))
   ((at ?a ?c1)(total-fuel-used ?total-fuel)(fuel ?a ?fuel))
   ((at ?a ?c2)(total-fuel-used ?new-total)(fuel ?a ?new-fuel))
 0.001)

(:operator (!!preprocessing ?problem-name)
           ((totaltime-coeff ?tc) (fuelused-coeff ?fc)
            (eval (setf *tc* ?tc))
            (eval (setf *fc* ?fc)))
           ()
           ()
     0)

(:operator (!!assert ?g )
        ()
        ()
         ?g
  0)
(:operator (!!ra ?D ?A )
        ()
        ?D
        ?A
 0)
```





```
;;;;
;;;;; Main Methods
;;;;;

 (:method (transport-person ?p ?c)
    Case1 ((at ?p ?c))
          ())

 (:method (transport-person ?p ?c2)
    Case2 (:sort-by ?num #'>
     ((at ?p ?c1)
            (at ?a ?c1)
            (aircraft ?a)
      (onboard ?a ?num)))
          ((!!assert ((dest ?a ?c1)))
            (:immediate board ?p ?a ?c1)
            (!!assert ((dest ?a ?c2)))
            (:immediate upper-move-aircraft-no-style ?a ?c2)
            (:immediate debark ?p ?a ?c2)))

 (:method (transport-person ?p ?c2)
    Case3 (:sort-by ?cost #'<
     ((at ?p ?c1)
            (aircraft ?a)
            (at ?a ?c3)
      (different ?c1 ?c3)
            (forall (?c) ((dest ?a ?c)) ((same ?c ?c1)))
            (imply ((different ?c3 ?c1))
            (not (possible-person-in ?c3)))
      (travel-cost-info ?a ?c3 ?c1 ?cost ?style)))
          ((!!assert ((dest ?a ?c1)))
            (:immediate upper-move-aircraft ?a ?c1 ?style)
            (:immediate board ?p ?a ?c1)
            (!!assert ((dest ?a ?c2)))
            (:immediate upper-move-aircraft-no-style ?a ?c2)
            (:immediate debark ?p ?a ?c2)))

 (:method (upper-move-aircraft ?a ?c ?style)
    Case1 ((at ?a ?c))
          ()
    Case2 ((at ?a ?somecity))
          ((move-aircraft ?a ?somecity ?c ?style)))

 (:method (upper-move-aircraft-no-style ?a ?c)
    Case1 ((at ?a ?c))
          ()
    Case2 (:sort-by ?cost #'<
           ((at ?a ?somecity)
             (travel-cost-info ?a ?somecity ?c ?cost ?style)))
          ((move-aircraft ?a ?somecity ?c ?style)))
```





```
(:- (travel-cost-info ?a ?from ?to ?cost slow)
    CASE1
     ((capacity ?a ?cap) (distance ?from ?to ?dist)
      (slow-burn ?a ?burn) (eval (< ?cap (* ?dist ?burn)))
      (assign ?cost most-positive-fixnum))

    CASE2
     ((distance ?from ?to ?dist) (fuel ?a ?fuel)
      (slow-burn ?a ?burn)
      (eval (>= ?fuel (* ?dist ?burn)))
      (assign ?cost (float (/
                            (+ *tc*
                               (* *fc*
                                  (* ?dist ?burn)))
                            1))))

   CASE3
     ((capacity ?a ?cap)(distance ?from ?to ?dist)
      (slow-burn ?a ?burn)
      (assign ?cost (float (/
                            (+ (* *tc* 2)
                               (* *fc*
                                  (* ?dist ?burn)))
                            1)))))
 (:- (travel-cost-info ?a ?from ?to ?cost fast)
    CASE1
     ((capacity ?a ?cap) (distance ?from ?to ?dist)
      (fast-burn ?a ?burn) (eval (< ?cap (* ?dist ?burn)))
      (assign ?cost most-positive-fixnum))
    CASE2
     ((distance ?from ?to ?dist) (fuel ?a ?fuel)
      (zoom-limit ?a ?limit) (onboard ?a ?num) (eval (< ?num ?limit))
      (fast-burn ?a ?burn)
      (eval (>= ?fuel (* ?dist ?burn)))
      (assign ?cost (float (/
                            (+ *tc*
                               (* *fc*
                                  (* ?dist ?burn)))
                            1))))
    CASE3
     ((capacity ?a ?cap)(distance ?from ?to ?dist)
      (fast-burn ?a ?burn)
      (zoom-limit ?a ?limit) (onboard ?a ?num) (eval (< ?num ?limit))
      (assign ?cost (float (/
                            (+ (* *tc* 2)
                               (* *fc*
                                  (* ?dist ?burn)))
                            1)))))
```





```
  (:method (move-aircraft ?a ?c1 ?c2 slow)
           ((fuel ?a ?fuel) (distance ?c1 ?c2 ?dist)
      (slow-burn ?a ?burn)
      (eval (> ?fuel (* ?dist ?burn))))
           ((fly ?a ?c1 ?c2))
     ()
     ((refuel ?a ?c1)
      (:immediate fly ?a ?c1 ?c2)))

  (:method (move-aircraft ?a ?c1 ?c2 fast)
           ((fuel ?a ?fuel) (distance ?c1 ?c2 ?dist)
      (fast-burn ?a ?burn)
      (eval (> ?fuel (* ?dist ?burn))))
            ((zoom ?a ?c1 ?c2))
     ()
     ((refuel ?a ?c1)
      (:immediate zoom ?a ?c1 ?c2)))

  (:method (transport-aircraft ?a ?c)
           ((not (no-use ?a)))
           ((!!assert ((no-use ?a)))
             (:immediate upper-move-aircraft-no-style ?a ?c)
             (:immediate !!ra ((no-use ?a)) ())))
```